\documentclass[runningheads]{llncs}

\usepackage{graphicx}

\usepackage{multirow}

\usepackage[T1]{fontenc}

\usepackage{caption}
\usepackage{subcaption}

\usepackage{comment}
\usepackage{amsmath,amssymb} 
\usepackage{color}
\usepackage{graphicx}
\usepackage{float}
\usepackage{booktabs}
\usepackage{tablefootnote}
\usepackage{longtable}
\usepackage{tabularx}
\usepackage{multirow}
\usepackage{boldline}
\usepackage{verbatim}
\usepackage{graphicx}
\usepackage{pifont}
\usepackage{tikz}
\usepackage[pagebackref,breaklinks,colorlinks]{hyperref}
\usepackage[capitalize]{cleveref}

\usepackage[accsupp]{axessibility}  

\usepackage{times}
\usepackage{epsfig}
\usepackage{graphicx}
\usepackage{amsmath}
\usepackage{amsmath, amssymb}

\usepackage{amssymb}
\usepackage{float}
\usepackage[numbers,sort,compress]{natbib}
\usepackage{graphicx,booktabs}
\usepackage{xcolor}
\usepackage{amsmath}
\usepackage{relsize}

\usepackage{xcolor}
\usepackage{graphicx}

\usepackage[font=scriptsize]{caption}

\usepackage{tkz-euclide}
\usepackage{amsmath}
\usepackage{amsfonts}
\usepackage{subcaption}
\usepackage{amssymb}
\usepackage{pifont}

\newcommand{\bigsum}[3]{\mathlarger{\sum}_{#1 = #2}^{#3}}

\usepackage[utf8]{inputenc}
\usepackage{amsmath, amssymb, latexsym}

\usepackage{xspace}
\usepackage{pgfplots,pgfplotstable}
\pgfplotsset{compat=1.7}
\usetikzlibrary{shapes.geometric}
\usepackage{hyperref}
\hypersetup{colorlinks,linkcolor={red},citecolor={green},urlcolor={red}} 

\hypersetup{pdfauthor={Name}}
\usetikzlibrary{decorations.pathreplacing}
\usepackage{adjustbox}
\usepackage{xcolor}

\restylefloat{table}

\usetikzlibrary{fit}

\def\checkmark{\tikz\fill[scale=0.4](0,.35) -- (.25,0) -- (1,.7) -- (.25,.15) -- cycle;}
\newcommand{\xmark}{\ding{55}}%

\usepackage{xcolor}

\definecolor{aogreen}{rgb}{0.0, 0.5, 0.0}

\renewcommand{\vec}[1]{\ensuremath{\mathbf{#1}}}
\newcommand{\mat}[1]{\ensuremath{\mathbf{#1}}}
\newcommand{\mob}{\ensuremath{\operatorname{M\ddot{o}bius}}}
\newcommand{\graphconv}{\ensuremath{\ast_\mathcal{G}}}
\newcommand{\transpose}{\ensuremath{^\top}}

\begin{document}

\pagestyle{headings}
\mainmatter

\newcommand{\expect}[1]{\ensuremath{\operatorname{\mathbb{E}}\!\left[ #1 \right]}}
\newcommand{\var}[1]{\ensuremath{\operatorname{Var}\!\left[ #1 \right]}}
\newcommand{\ie}[0]{\emph{i.e. }}
\newcommand{\eg}[0]{\emph{e.g. }}

\title{3D Human Pose Estimation\\Using M\"obius Graph Convolutional Networks} %

\titlerunning{M\"obiusGCN}
\author{Niloofar Azizi\inst{1} \and
Horst Possegger\inst{1} \and
Emanuele Rodol\`a\inst{2} \and
Horst Bischof\inst{1}}
\authorrunning{Azizi et al.}

\institute{Graz University of Technology, Graz, Austria
\email{\{azizi, possegger, bischof\}@icg.tugraz.at}\\
 \and
 Sapienza University of Rome, Rome, Italy\\
\email{rodola@di.uniroma1.it}}

\maketitle
\begin{abstract}
3D human pose estimation is fundamental to understanding human behavior. Recently, promising results have been achieved by graph convolutional networks (GCNs), which achieve state-of-the-art performance and provide rather light-weight architectures.
However, a major limitation of GCNs is their inability to encode all the transformations between joints explicitly. To address this issue, we propose a novel spectral GCN using the M\"obius transformation (M\"obiusGCN). In particular, this allows us to directly and explicitly encode the transformation between joints, resulting in a significantly more compact representation. Compared to even the lightest architectures so far, our novel approach requires $90$--$98\%$ fewer parameters, \ie our lightest M\"obiusGCN uses only $0.042\text{M}$ trainable parameters. Besides the drastic parameter reduction, explicitly encoding the transformation of joints also enables us to achieve state-of-the-art results.
We evaluate our approach on the two challenging pose estimation benchmarks, Human3.6M and MPI-INF-3DHP, demonstrating both state-of-the-art results and the generalization capabilities of M\"obiusGCN.
\end{abstract}

\section{Introduction}
Estimating 3D human pose helps to analyze human motion and behavior, thus enabling high-level computer vision tasks such as action recognition~\cite{luvizon20182d}, sports analysis~\cite{wang2019ai, rematas2018soccer},  augmented and virtual reality~\cite{han2018viton}. Although human pose estimation approaches already achieve impressive results in 2D, this is not sufficient for many analysis tasks, because several 3D poses can project to exactly the same 2D pose. Thus, knowledge of the third dimension can significantly improve the results on the high-level tasks.

Estimating 3D human joint positions, however, is  challenging. On the one hand, there are only very few labeled datasets because 3D annotations are expensive. On the other hand, there are self-occlusions, complex joint inter-dependencies, small and barely visible joints, changes in appearance like clothing and lighting, and the many degrees of freedom of the human body.

To solve 3D human pose estimation, some methods utilize multi-views~\cite{rhodin2018unsupervised, zhang2021adafuse}, synthetic datasets~\cite{peng2018jointly}, or motion~\cite{li2021lifting, shere2021temporally}.  For improved generalization, however, we follow the most common line of work and estimate 3D poses given only the 2D estimate of a single RGB image as input, similar to \cite{martinez2017simple, liu2020comprehensive, pavllo20193d}.
First, we compute 2D pose joints given RGB images using an off-the-shelf architecture. Second, we approximate the 3D pose of the human body using the estimated 2D joints.

\begin{figure}
\centering
\scalebox{.7}{
\resizebox{\linewidth}{!}{
\tikzset{
        path00 image/.style={
                path picture={
                        \node at (path picture bounding box.center) {
                                \includegraphics[width=7cm]{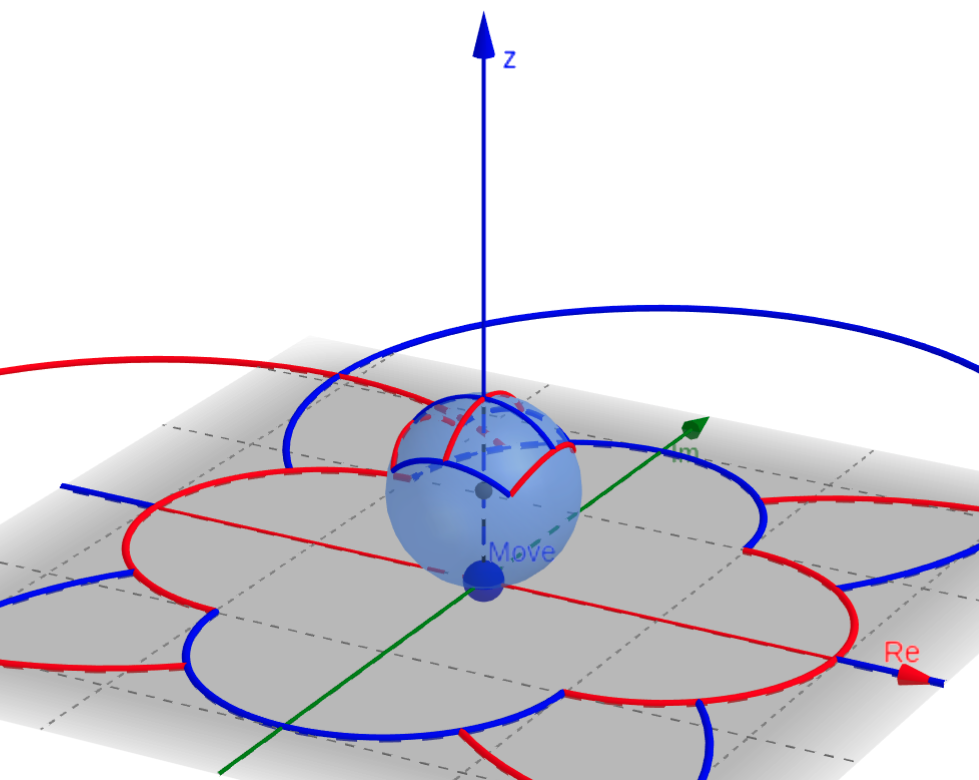}};}},
}
\tikzset{
        path01 image/.style={
                path picture={
                        \node at (path picture bounding box.center) {
                                \includegraphics[width=7cm]{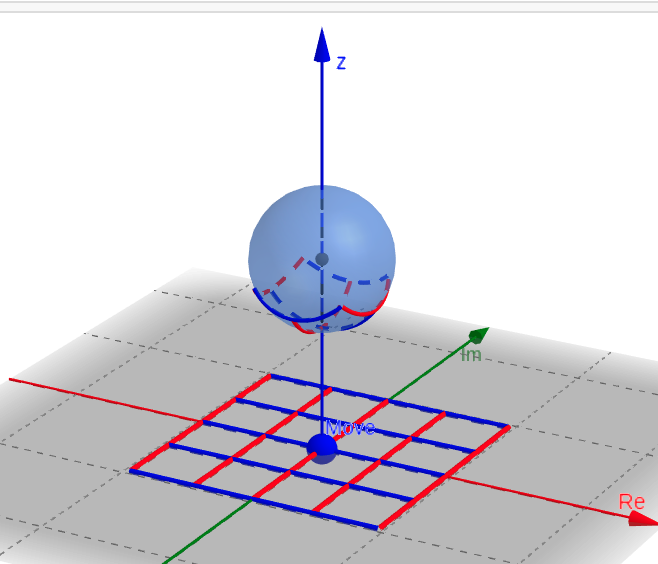}};}},
}

\begin{tikzpicture}[node distance=0mm, font=\sffamily, x=1cm, y=1cm]
        
        \node (pose) at (0,0) [inner sep=0pt, anchor=center] 
        {\includegraphics[width=10cm]{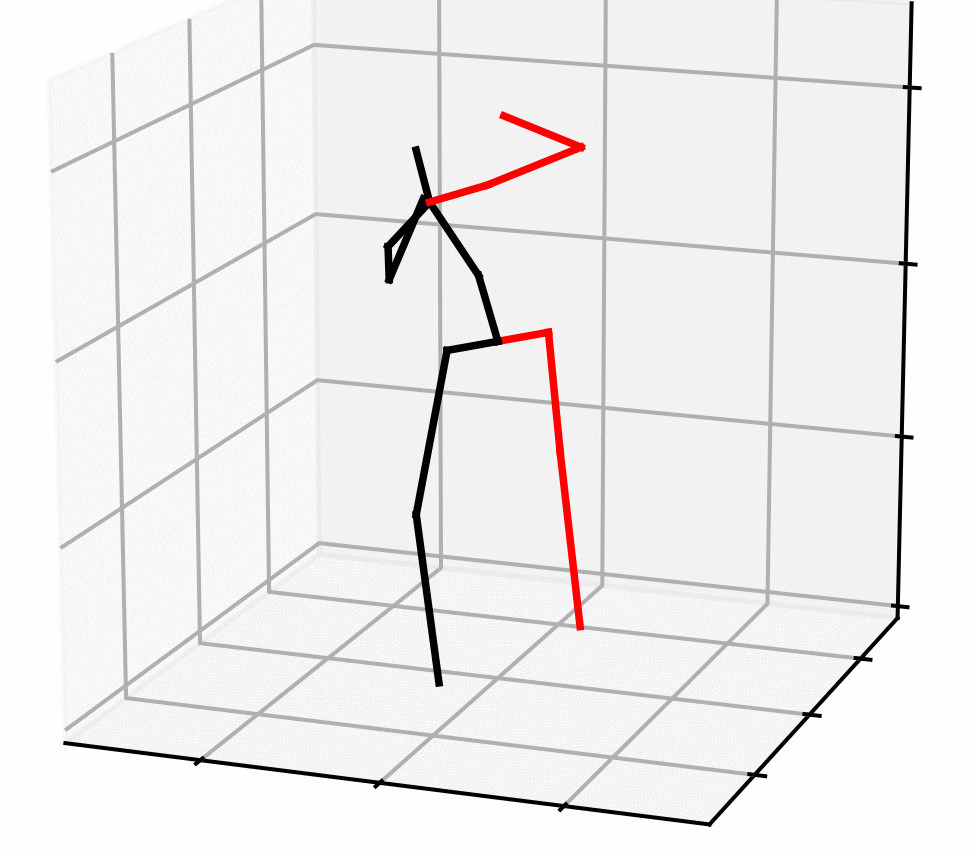}};
        \node (input) at (-6,-2.5) [inner sep=0pt, anchor=center] {\includegraphics[width=5.cm]{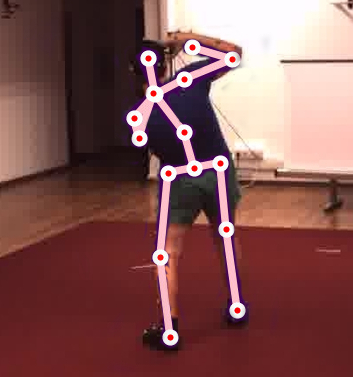}};
        \draw [path00 image, draw=none] (-8,4) circle (3);
        \node (c0l) at (-8, 4) [circle, minimum width = 5.98cm, name path=circle]{};
        \node (c0) at (-8, 4) [circle, draw, minimum width = 6cm, line width=.7mm, color= gray!60!black, name path=circle]{};

        \draw [path01 image, draw=none] (8,-1) circle (3);
        \node (c1l) at (8, -1) [circle, minimum width = 5.98cm, name path=circle]{};
        \node (c1) at (8, -1) [circle, draw, minimum width = 6cm, line width=.7mm, color= gray!60!black, name path=circle]{};

        \node (l01) at (-1.1, 1.9) {};
        \draw[color = gray!60!black, line width=.5mm, dashed] (l01)  -- (tangent cs:node=c0l,point={(l01)},solution=1) coordinate[];
        \draw[color = gray!60!black, line width=.5mm, dashed] (l01)  -- (tangent cs:node=c0l,point={(l01)},solution=2) coordinate[];
        \node (l01) at (0.8, 0.9) {};
        \draw[color = gray!60!black, line width=.7mm, dashed] (l01)  -- (tangent cs:node=c1l,point={(l01)},solution=1) coordinate[];
        \draw[color = gray!60!black, line width=.7mm, dashed] (l01)  -- (tangent cs:node=c1l,point={(l01)},solution=2) coordinate[];

        \node (txtpose) [above=.6cm of pose] {\Huge\textcolor{black}{Estimated 3D Pose}};
        \node (txtinput) [below=.1cm of input] {\LARGE\textcolor{black}{\hspace{1.5cm}Input}};
        \node (txtmobius2a) [below=0.25cm of c1] {\LARGE \textcolor{black}{M\"obius Transformation}};
\end{tikzpicture}}}
\caption{Our M\"obiusGCN accurately learns the transformation (particularly the rotation) between joints by leveraging the M\"obius transformation given estimated 2D joint positions from a single RGB image.}
\label{fig1}
\end{figure}
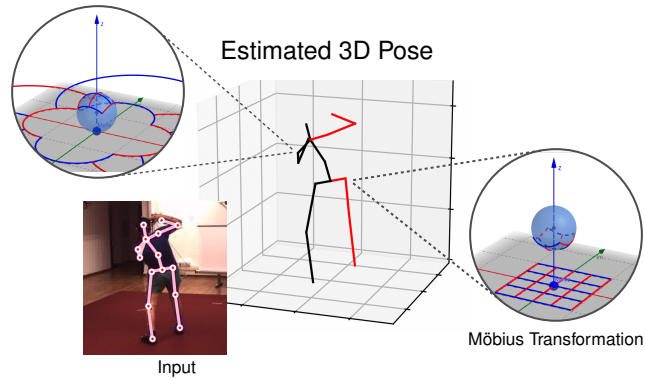
With the advent of deep learning methods, the accuracy of 3D human pose estimation has significantly improved, \eg~\cite{martinez2017simple, pavllo20193d}. Initially, these improvements were driven by CNNs (Convolutional Neural Networks). 
However, these assume that the input data is stationary, hierarchical, has a grid-like structure, and shares local features across the data domain. The convolution operator in the CNN assumes that the nodes have fixed neighbor positions and a fixed number of neighbor nodes. Therefore, CNNs are not applicable to graph-structured data. The input to 3D pose estimation from 2D joint positions, however, is graph-structured data. Thus, to handle this irregular nature of the data, GCNs  (Graph Convolutional Networks) have been proposed \cite{bronstein2017geometric}.

GCNs are able to achieve state-of-the-art performance for 2D-to-3D human pose estimation with comparably few parameters, \eg~\cite{zhaoCVPR19semantic}. Nevertheless, to the best of our knowledge, none of the previous GCN approaches explicitly models the inter-segmental angles between joints. 
Learning the inter-segmental angle distribution explicitly along with the translation distribution, however, leads to encoding better feature representations.
Thus, we present a novel spectral GCN architecture, \emph{M\"obiusGCN}, to accurately learn the transformation between joints and to predict 3D human poses given 2D joint positions from a single RGB image. To this end, we leverage the M\"obius transformation on the eigenvalue matrix of the graph Laplacian. Previous GCNs applied for estimating the 3D pose of the human body are defined in the real domain, \eg~\cite{zhaoCVPR19semantic}.
Our M\"obiusGCN operates in the complex domain, which allows us to encode all the transformations (\ie inter-segmental angles and translation) between nodes simultaneously (Figure \ref{fig1}). 

An enriched feature representation achieved by encoding the transformation distribution between joints using a M\"obius transformation provides us with a compact model. A light DNN architecture makes the network independent of expensive hardware setup, enabling the use of mobile phones and embedded devices at inference time. This can be achieved by our compact M\"obiusGCN architecture.

Due to a large number of weights that need to be estimated, fully-supervised state-of-the-art approaches need an enormous amount of annotated data, where data annotation is both time-consuming and requires expensive setup. Our M\"obiusGCN, on the contrary, requires only a tiny fraction of the model parameters, which allows us to achieve competitive results with significantly fewer annotated data. 

We summarize our main contributions as follows:
\begin{itemize}
\itemsep0em
    \item We introduce a novel spectral GCN architecture leveraging the Möbius transformation to explicitly encode the pose, in terms of inter-segmental angles and translations between joints.
    \item We achieve state-of-the-art 3D human pose estimation results, despite requiring only a fraction of the model parameters (\ie 2--9\% of even the currently lightest approaches).
    \item Our light-weight architecture and the explicit encoding of transformations lead to state-of-the-art performance compared to other semi-supervised methods, by training only on a reduced dataset given estimated 2D human joint positions.
\end{itemize}

\section{Related Work}

\textbf{Human 3D Pose Estimation.} The classical approaches addressing the 3D human pose estimation task are usually based on hand-engineered features and leverage prior assumptions, \eg using motion models~\cite{sminchisescu20083d} or other common heuristics~\cite{h36m_pami, ramakrishna2012reconstructing}. Despite good results, their major downside is the lack of generality.

Current state-of-the-art approaches in computer vision, including 3D human pose estimation, are typically based on DNNs (Deep Neural Networks), \eg~\cite{sarandi2020metrabs, luo2021multi, ma2021context, li2019generating}. To use these architectures, it is assumed that the statistical properties of the input data have locality, stationarity, and multi-scalability~\cite{henaff2015deep}, which reduces the number of parameters.

Although DNNs achieve state-of-the-art in spaces governed by Euclidean geometry, a lot of the real-world problems are of a non-Euclidean nature. For these problem classes, GCNs have been introduced. There are two types of GCNs: spectral GCN and spatial GCN. Spectral GCNs rely on the Graph Fourier Transform, which analyzes the graph signals in the vector space of the graph Laplacian matrix. The second category, spatial GCN, is based on feature transformations and neighborhood aggregation on the graph. Well-known spatial GCN approaches include Message Passing Neural Networks~\cite{gilmer2017neural} and GraphSAGE~\cite{hamilton2017inductive}. 

For 3D human pose estimation, GCNs achieve competitive results with comparably few parameters. Pose estimation with GCNs has been addressed, \eg in~\cite{liu2020comprehensive, xu2021graph,zhaoCVPR19semantic}.~\citet{xu2021graph} proposed Graph Stacked Hourglass Networks (GraphSH), in which graph-structured features are processed across different scales of human skeletal representations. \citet{liu2020comprehensive} investigated different combinations of feature transformations and neighborhood aggregation of spatial features.
They also showed the benefits of using separate weights to incorporate a node's self-information.
~\citet{zhaoCVPR19semantic} proposed semantic GCN (SemGCN), which currently represents the lightest architecture ($0.43\text{M}$). The key idea is to learn the adjacency matrix, which lets the architecture encode the graph's semantic relationships between nodes. In contrast to SemGCN, we can further reduce the number of parameters by an order of magnitude ($0.042\text{M}$) by explicitly encoding the transformation between joints. The key ingredient to this significant reduction is the M\"obius transformation.

\textbf{M\"{o}bius Transformation.} The M\"{o}bius transformation has been used in neural networks as an activation function~\cite{ozdemir2011complex,mandic2009complex}, in hyperbolic neural networks~\cite{NEURIPS2018_dbab2adc}, for data augmentation~\cite{zhou2021data}, and for knowledge graph embedding~\cite{nayyeri20205}.
Our work is the first to introduce M\"{o}bius transformations for spectral graph convolutional networks.
To utilize the M\"{o}bius transformation, we have to design our neural network in the complex domain. The use of complex numbers (analysis in polar coordinates) to harness phase information along with the signal amplitude is well established in signal processing~\cite{mandic2009complex}. By applying the M\"obius transformation, we let the architecture encode the transformations (\ie~inter-segmental angle and translation) between joints explicitly, which leads to a very compact architecture.

\textbf{Handling Rotations.} Learning the rotation between joints in skeletons has been investigated previously for 3D human pose estimation~\cite{zhou2016deep,parameswaran2004view,barron2001estimating}. Learning the rotation using Euler angles or quaternions, however, has obvious issues like discontinuities~\cite{saxena2009learning, zhou2019continuity}. 
Continuous functions are easier to learn in neural networks~\cite{zhou2019continuity}.
~\citet{zhou2019continuity} tackle the discontinuity by lifting the problem to 5 and 6 dimensions. Another direction of research focuses on designing DNNs for inverse kinematics with the restricted assumption of putting joint angles in a specific range to avoid discontinuities. However, learning the full range of rotations is necessary for many real-world problems~\cite{zhou2019continuity}. 
Our M\"obiusGCN is continuous by definition and thus, allows us to elegantly encode rotations.

\textbf{Data Reduction.} 
 A major benefit of light architectures is that they require smaller datasets to train.
Semi-supervised methods require a small subset of annotated data and a large set of unannotated data for training the network. These methods are actively investigated in different domains, considering the difficulty of providing annotated datasets.
Several semi-supervised approaches for 3D human pose estimation benefit from applying more constraints over the possible space solutions by utilizing multi-view approaches using RGB images from different cameras~\cite{rhodin2018unsupervised, rhodin2018learning, yao2019monet, wandt2021canonpose, mitra2020multiview}. 
These methods need expensive laboratory setups to collect synchronized multi-view data.

~\citet{pavllo20193d} phrase the loss over the back-projected estimated 3D human pose to 2D human pose space conditioned on time. 
~\citet{tung2017adversarial} use generative adversarial networks to reduce the required annotated data for training the architecture. 
~\citet{iqbal2020weakly} relax the constraints using weak supervision; they introduce an end-to-end architecture that estimates 2D pose and depth independently, and uses a consistency loss to estimate the pose in 3D.

Our compact M\"obiusGCN achieves competitive state-of-the-art results with only scarce training samples. M\"obiusGCN does not require any multi-view setup or temporal information. Further, it does not rely on large unlabeled datasets. It just requires a small annotated dataset to train. In contrast, the previous semi-supervised methods require complicated architectures and a considerable amount of unlabeled data during the training phase. 

\section{Spectral Graph Convolutional Network}
\tikzset{arrow/.style={-stealth, thick, draw=gray!80!black}}
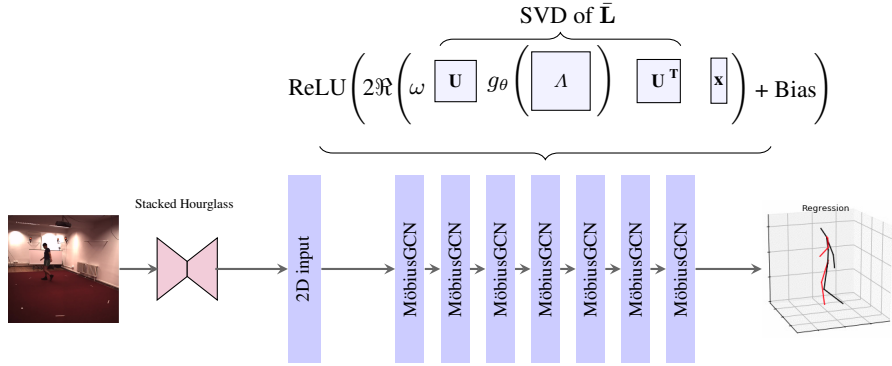
\begin{figure*}
\centering
  \hspace*{0.01px}
\begin{tikzpicture}[outer/.style={},
square/.style={draw, solid, fill=blue!5,regular polygon,regular polygon sides=4}]
	\node[inner sep=0pt] (X) at (-3.2,0)
    {\includegraphics[width=0.12\textwidth]{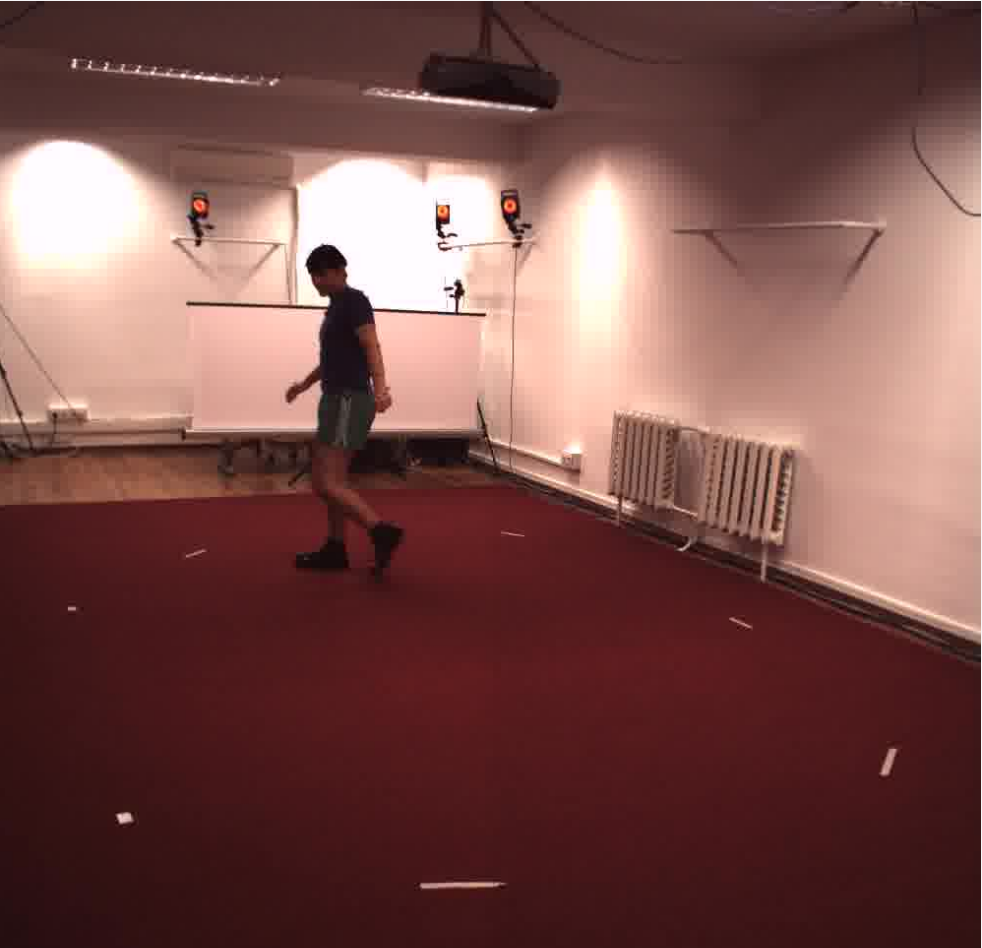}};
	\draw[fill=purple!20] ([xshift=0.5cm,yshift=-0.3cm]X.north east) -- ([xshift=0.9cm,yshift=0.1cm]X.east) -- ([xshift=0.9cm,yshift=-0.1cm]X.east) -- ([xshift=0.5cm,yshift=0.3cm]X.south east) -- cycle; 
	
	\draw[fill=purple!20] ([xshift=0.9cm,yshift=0.1cm]X.east) --
	([xshift=1.3cm,yshift=-0.3cm]X.north east) -- 
	([xshift=1.3cm,yshift=0.3cm]X.south east) -- 
	([xshift=0.9cm,yshift=-0.1cm]X.east) -- 	cycle;
	\node at (-1.6,0.85) {{\tiny Stacked Hourglass}};
	\node at (-1.83,0) (temp){};
	\node at (-1.3,0) (temp1){};
    \node[fill=blue!20, minimum width=2.5cm, minimum height=0.25cm, rotate=90] (MobiusGCN1) at (1.4,0) {{\scriptsize M\"obiusGCN}};
    \node[fill=blue!20, minimum width=2.5cm, minimum height=0.25cm, rotate=90] (relu1) at (2.,0) {{\scriptsize M\"obiusGCN}};
    \node[fill=blue!20, minimum width=2.5cm, minimum height=0.25cm, rotate=90] (MobiusGCN2) at (2.6,0) {{\scriptsize M\"obiusGCN}};
    \node[fill=blue!20, minimum width=2.5cm, minimum height=0.25cm, rotate=90] (relu2) at (3.2,0) {{\scriptsize M\"obiusGCN}};
    \node[fill=blue!20, minimum width=2.5cm, minimum height=0.25cm, rotate=90] (MobiusGCN3) at (3.8,0) {{\scriptsize M\"obiusGCN}};
    \node[fill=blue!20, minimum width=2.5cm, minimum height=0.25cm, rotate=90] (relu3) at (4.4,0) {{\scriptsize M\"obiusGCN}};
    \node[fill=blue!20, minimum width=2.5cm, minimum height=0.25cm, rotate=90] (MobiusGCN4) at (5.,0) {{\scriptsize M\"obiusGCN}};
    \node[fill=blue!20, minimum width=2.5cm, minimum height=0.25cm, rotate=90] (twod) at (0.0,0) {{\scriptsize 2D input}};

    \node[inner sep=0pt] (3d) at (7.,0)
    {\includegraphics[width=0.15\textwidth]{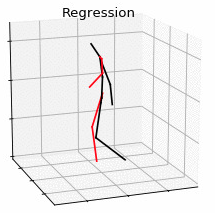}};

\node[outer] (A) at (3.7,2.5) {
    \begin{tikzpicture}[node distance=1cm,outer sep = 0pt]
        \node [square,minimum width=0.4cm] (u) {\scriptsize$\textbf{U}$};
        \node [square,minimum width=1.1cm] at (1.4,0) (lambda) {\scriptsize \textbf{$\Lambda$}};
        \node [square,minimum width=0.4cm] at (2.7,0) (ut) {\scriptsize$\textbf{U}^{}$};
        \node at (2.9,0.1) (t) {\tiny$\textbf{T}$};
        \coordinate (dm1) at (3.5,0.2);
        \coordinate (dm2) at (3.5,-0.19);
        \node[rectangle,draw,solid, fill=blue!5,minimum width=0.1cm] [fit = (dm1) (dm2)] (bx4) {};
        \node[align=center,font=\large] at (bx4.center) {$\scriptsize \textbf{x}$};
        \node at (0.7,0) {{$g_{\theta}$\Bigg(}};
        \node at (2,0) {{\Bigg)}};
    \end{tikzpicture}
                };
    \node at (0.3,2.4) {{\text{ReLU}\Bigg (}};
    \node at (1.5,2.4) {$\omega$};
    \node at (1.1,2.4) {{2$\Re$\Bigg(}};
    \node at (5.8,2.4) {{\Bigg)}};
    \node at (6.4,2.4) {{+ \text{Bias}}};
    \node at (6.9,2.4) {{\Bigg)}};
    \draw [arrow] (X) -- (temp);
    \draw [arrow] (temp1) -- (twod);
    \draw [arrow] (twod) -- (MobiusGCN1);
    \draw [arrow] (MobiusGCN1) -- (relu1);
    \draw [arrow] (relu1) -- (MobiusGCN2);
    \draw [arrow] (MobiusGCN2) -- (relu2);
    \draw [arrow] (relu2) -- (MobiusGCN3);
    \draw [arrow] (MobiusGCN3) -- (relu3);
    \draw [arrow] (relu3) -- (MobiusGCN4);
    \draw [arrow] (MobiusGCN4) -- (3d);
    \draw [decorate,decoration={brace,amplitude=5pt,mirror,raise=4ex}]
  (0.2,2.2) -- (6.2,2.2) node[midway,yshift=-3em]{};

    \node at (3.5,3.4) {{SVD of $\Bar{\mat{L}}$}};
    \draw [decorate,decoration={brace,amplitude=5pt,raise=4ex}]
  (1.8,2.4) -- (5.,2.4) node[midway,yshift=-3em]{};
    
\end{tikzpicture}
  \vskip 6px
  \caption{The complete pipeline of the proposed M\"obiusGCN architecture; The output of the off-the-shelf stacked hourglass architecture~\cite{newell2016stacked}, \ie~estimated 2D joints of the human body, is the input to the M\"obiusGCN architecture. The M\"obiusGCN architecture locally encodes the transformation between the joints of the human body. SVD is the singular value decomposition of the normalized Laplacian matrix. Function $g_{\theta}$ is the M\"obius transformation applied on the eigenvalues of the eigenvalue matrix independently. \textbf{x} is the graph signal and $\omega$ are the learnable parameters, both in the complex domain.}
  \label{figure:hourglass}
\end{figure*}
\vspace{-30px}
\subsection{Graph Definitions}
Let $\mathcal{G}(V, E)$ represent a graph consisting of a finite set of $N$ vertices, $V =\{\upsilon_1, \dots, 	\upsilon_N \}$, and a set of $M$ edges $E  =\{e_1, \dots, 	e_M \}$, with $e_j = (\upsilon_{i}, \upsilon_{k})$ where $\upsilon_{i}, \upsilon_{k} \in V$.
The graph's adjacency matrix $\mathbf{A}_{N \times N}$ contains $1$ in case two vertices are connected and $0$ otherwise. 
$\mathbf{D}_{N \times N}$ is a diagonal matrix where $\mathbf{D}_{ii}$ is the degree of vertex $\upsilon_{i}$.
A graph is directed if $(\upsilon_i, \upsilon_k) \neq (\upsilon_k, \upsilon_i)$, otherwise it is an undirected graph. For an undirected graph, the adjacency matrix is symmetric.
The non-normalized graph Laplacian matrix is defined as $\mat{L} = \mat{D} - \mat{A},$
and can be normalized to $\Bar{\mat{L}} = \mat{I} - \mat{D}^{-\frac{1}{2}}\mat{A}\mat{D}^{-\frac{1}{2}},$
where $\mathbf{I}$ is the identity matrix. 
$\Bar{\mathbf{L}}$ is  real, symmetric, and positive semi-definite. Therefore, it has $N$ ordered, real, and non-negative eigenvalues $\{\lambda_i : i = 1, \dots , N \}$ and corresponding orthonormal eigenvectors $\{\mathbf{u}_i : i = 1, \dots , N \}$. 

A signal $\mathbf{x}$ defined on the nodes of the graph is a vector $\mathbf{x} \in \mathbb{R}^N$, where its $i$-th component represents the function value at the $i$-th vertex in $V$. Similarly, $\mathbf{X} \in \mathbb{R}^{N \times d}$ is called a $d$-dimensional graph signal on $\mathcal{G}$~\cite{shuman2013emerging}. 
\subsection{Graph Fourier Transform}
Graph signals $\textstyle\mathbf{x} \in \mathbb{R}^N$ admit a graph Fourier expansion $\mathbf{x}=\sum_{i=1}^N \langle \mathbf{u}_{i},\mathbf{x}\rangle \mathbf{u}_i$, where $\mathbf{u}_i, i=1,\dots,N$ are the eigenvectors of the  graph Laplacian~\cite{shuman2013emerging}. Eigenvalues and eigenvectors of the graph Laplacian matrix are analogous to frequencies and sinusoidal basis functions in the classical Fourier series expansion. 
\subsection{Spectral Graph Convolutional Network}
Spectral GCNs~\cite{bruna2013spectral} build upon the  graph Fourier transform. Let $\vec{x}$ be the graph signal and $\vec{y}$ be the graph filter on graph $\mathcal{G}$. The graph convolution $\ast_\mathcal{G}$ can be defined as
\begin{equation}
    \vec{x} \ast_\mathcal{G} \vec{y} = \mat{U}(\mat{U}\transpose\vec{x} \odot \mat{U}\transpose\vec{y}) 
    \label{naiveGCN}
\end{equation}
where the matrix $\mat{U}$ contains the eigenvectors of the normalized graph Laplacian and $\odot$ is the Hadamard product.
This can also be written as
 \begin{equation}
    \vec{x} \;\ast_\mathcal{G} \; g_\theta = \mat{U}g_{\theta}(\mat{\Lambda})\mat{U}\transpose\vec{x},
    \label{eq-naiveGCN}
\end{equation}
where $g_{\theta}(\mat{\Lambda})$ is a diagonal matrix with the parameter $\theta \in \mathbb{R}^N$ as a vector of Fourier coefficients.

\subsection{Spectral Graph Filter}
Based on the corresponding definition of $g_\theta$ in Eq.~\eqref{eq-naiveGCN}, spectral GCNs can be classified into spectral graph filters with smooth functions and spectral graph filters with rational functions.

\textbf{Spectral Graph Filter with Smooth Functions.}~\citet{henaff2015deep} proposed defining $g_{\theta}(\boldsymbol{\Lambda})$ to be a smooth function (smoothness in the frequency domain corresponds to the spatial decay), to address the localization problem.

 ~\citet{defferrard2016convolutional} proposed defining the function $g_{\theta}$ in such a way to be directly applicable over the Laplacian matrix to address the computationally costly Laplacian matrix decomposition and multiplication with the eigenvector matrix in Eq.~\eqref{eq-naiveGCN}.

~\citet{kipf2016semi} defined $g_{\theta}(\mathbf{L})$ to be the Chebychev polynomial by assuming all the eigenvalues in the range of $[-1, 1]$. Computing the polynomials of the Chebychev polynomial, however, is computationally expensive. Also, considering polynomials with higher orders causes overfitting. Therefore,~\citet{kipf2016semi} approximated the Chebychev polynomial with its first two orders.

\textbf{Spectral Graph Filter with Rational Functions.} Fractional spectral GCNs, unlike polynomial spectral GCNs, can model sharp changes in the frequency response~\cite{bianchi2021graph}.
\citet{levie2017cayleynets} put the eigenvalues of the Laplacian matrix on the unit circle by applying the Cayley transform on the Laplacian matrix with a learned parameter, named spectral coefficient, that lets the network focus on the most useful frequencies.

Our proposed M\"obiusGCN is also a fractional GCN which applies the M\"obius transformation on the eigenvalue matrix of the normalized Laplacian matrix to encode the transformations between joints.

\section{M\"{o}biusGCN}
A major drawback of previous spectral GCNs is that they do not encode the transformation distribution between nodes explicitly. We address this by applying the M\"obius transformation function over the eigenvalue matrix of the decomposed Laplacian matrix. This simultaneous encoding of the rotation and translation distribution in the complex domain leads to better feature representations and fewer parameters in the network.

The input to our first block of M\"obiusGCN are the joint positions in 2D Euclidean space, given as $\mathcal{J} = \{J_i \in \mathbb{R} ^2| i = 1, \dots, \kappa\}$, which can be computed directly from the image. Our goal is then to predict the corresponding 3D Euclidean joint positions $\hat{\mathcal{Y}} = \{\hat{\mathcal{Y}}_i \in \mathbb{R}^3| i = 1, \dots, \kappa\}$.

We leverage the structure of the input data, which can be represented by a connected, undirected and unweighted graph.
The input graphs are fixed and share the same topological structure, which means the graph structure does not change, and each training and test example differs only in having different features at the vertices. In contrast to pose estimation, tasks like protein-protein interaction~\cite{velivckovic2017graph} are not suitable for our M\"obiusGCN, because there the topological structure of the input data can change across samples.

\subsection{M\"{o}bius Transformation}
The general form of a M\"{o}bius transformation~\cite{mandic2009complex} is given by
$f(z) = \frac{az+b}{cz+d}$ where $a, b, c, d, z \in \mathbb{C}$ satisfy $ad - bc \ne 0$. The M\"{o}bius transformation can be expressed as the composition of simple transformations.
Specifically, if $ c\neq 0,$ then:
\begin{itemize}
\itemsep0em
    \item $f_{1}(z)=z+d/c$ defines translation by $d/c$,
    \item $ f_{2}(z)=1/z$  defines inversion and reflection with respect to the real axis,
    \item $f_{3}(z)={\frac {bc-ad}{c^{2}}}z$ defines homothety and rotation,
    \item $f_{4}(z)=z+a/c$ defines the translation by $a/c$.
\end{itemize}
These functions can be composed to form the M\"obius transformation
\begin{equation*}
f(z) = f_{4}\circ f_{3}\circ f_{2}\circ f_{1}(z) = {\frac {az+b}{cz+d}}\,,
\end{equation*}
where $\circ$ denotes the composition of two functions $f$ and $g$ as
\begin{equation*}
    f \circ g(z) = f(g(z)).
\end{equation*}

 The M\"obius transformation is analytic everywhere except at the pole $z = -\frac{d}{c}$. 
 Since a M\"{o}bius transformation remains unchanged by scaling with a coefficient~\cite{mandic2009complex}, we normalize it to yield the determinant 1. 
We observed that in our gradient-based optimization setup, the M\"{o}bius transformation in each node converges into the fixed points. In particular, the M\"obius transformation can have two fixed points (loxodromic), one fixed point (parabolic or circular), or no fixed point. The fixed points can be computed by solving $\frac{az + b}{cz + d} = z$, which gives 
\begin{equation*}
    \gamma_{1,2} = \frac{a - d + \sqrt{(a-d)^2 - 4bc}}{2c}.
\end{equation*}

\subsection{M\"obiusGCN}

To predict the 3D human pose, we explicitly encode the local transformations between joints, where each joint corresponds to a node in the graph. To do so, we define $g_{\theta}(\mat{\Lambda})$ in Eq.~\eqref{eq-naiveGCN} to be the M\"{o}bius transformation applied to the Laplacian eigenvalues, resulting in the following fractional spectral graph convolutional network
\begin{equation}
    \vec{x} \graphconv g_\theta(\mat{\Lambda}) = \mat{U}\  \mob(\mat{\Lambda})\ \mat{U}\transpose\vec{x} = \bigsum{i}{0}{N - 1} \mob_i(\lambda_i)\vec{u}_i\vec{u}_i\transpose\vec{x}\,,
    \label{mobgcn--better-use-of-latex}
\end{equation}
where 
\begin{equation}
\mob_i(\lambda_i) = \frac{a_i\lambda_i+b_i}{c_i\lambda_i+d_i},    
\end{equation}
 with $a_i, b_i, c_i, d_i, \lambda_i \in\mathbb{C}$.

Applying the M\"obius transformation over the Laplacian matrix places the signal in the complex domain. To return back to the real domain, we sum it up with its conjugate
\begin{equation}
    \mat{Z} = 2\Re \{w\ \mat{U}\ \mob (\mat{\Lambda})\ \mat{U}\transpose \vec{x}\},
    \label{realify}
\end{equation}
where $w$ is the shared complex-valued learnable weight to encode different transformation features. This causes the number of learned parameters to be reduced by a factor equal to the number of joints (nodes of the graph). The inter-segmental angles between joints are encoded by learning the rotation functions between neighboring nodes. 

We can easily generalize this definition to the graph signal matrix $\mathbf{X} \in \mathbb{C}^{N\times d}$ with $d$ input channels (\ie a $d$-dimensional feature vector for every node) and $\mathbf{W} \in \mathbb{C}^{d \times F}$ feature maps. 
This defines a M\"obiusGCN block
\begin{equation}
    \mat{Z} = \sigma(2\Re \{\mat{U} \mob(\mat{\Lambda})\mat{U}\transpose\mat{X}\mat{W}\} + \vec{b}),
\label{block}
\end{equation}
where $\mathbf{Z} \in \mathbb{R}^{N\times F}$ is the convolved signal matrix,
$\sigma$ is a nonlinearity  (\eg \text{ReLU}~\cite{nair2010rectified}), and $\vec{b}$ is a bias term.

To encode enriched and generalized joint transformation feature representations, we make the architecture deep by stacking several blocks of M\"obiusGCN. Stacking these blocks yields our complete architecture for 3D pose estimation, as shown in Figure~\ref{figure:hourglass}.

To apply the M\"obius transformation over the matrix of eigenvalues of the Laplacian matrix, we encode the weights of the M\"obius transformation for each eigenvalue in four diagonal matrices $ \mat{A}, \mat{B}, \mat{C}, \mat{D}$ and compute
\begin{equation}
 \mat{U} \mob(\mat{\Lambda})\mat{U}\transpose = \mat{U}(\mat{A}\mat{\Lambda} + \mat{B})(\mat{C}\mat{\Lambda} + \mat{D})^{-1}\mat{U}\transpose.   
\end{equation}

\subsection{Why M\"obiusGCN is a Light Architecture}
As a direct consequence of applying the M\"obius transformation for the graph filters in polar coordinates, the filters in each block can encode the inter-segmental angle features between joints in addition to the translation features explicitly. By applying the M\"obius transformation, the graph filter scales and rotates the eigenvectors of the Laplacian matrix in the graph Fourier transform simultaneously. This leads to learning better feature representations and thus, yields a more compact architecture.

For a better understanding, consider the following analogy with the classical Fourier transform: While it can be hard to construct an arbitrary signal by a linear combination of basis functions with real coefficients (i.e., the signal is built just by changing the amplitudes of the basis functions), it is significantly easier to build a signal by using {\em complex} coefficients, which change both the phase and amplitude.

In previous spectral GCNs, specifically~\cite{kipf2016semi}, the Chebychev polynomials are only able to scale the eigenvectors of the Laplacian matrix, in turn requiring both more parameters and additional nonlinearities to encode the rotation distribution between joints implicitly. 

\subsection{Discontinuity}
Our model encodes the transformation between joints in the complex domain by learning the parameters of the normalized M\"obius transformation.
In the definition of the M\"obius transformation, if $ad - bc \neq 0$, then the M\"obius transformation is an injective function and thus, continuous by the definition of continuity for neural networks given by~\cite{zhou2019continuity}.
M\"obiusGCN does not suffer from discontinuities in representing inter-segmental angles, in contrast to Euler angles or quaternions. Additionally, this leads to significantly fewer parameters in our architecture.

\begin{table*}
\centering
\resizebox{12CM}{!}{%
 \begin{tabular}{c| c c c c c c c c c c c c c c c c c} 
 \hline
 \textbf{Protocol~\#1} &\text{\# Param.} & \text{Dir.} & \text{Disc.} & \text{Eat} & \text{Greet} & \text{Phone} & \text{Photo} & \text{Pose} & \text{Purch.} & \text{Sit} & \text{SitD.} & \text{Smoke} & \text{Wait} & \text{WalkD.} & \text{Walk} & \text{WalkT.} & \textbf{Average} \\ [0.5ex] 
 \hline
  ~\citet{martinez2017simple}\hypersetup{citecolor=black}   & 4.2M & 51.8 & 56.2 & 58.1 & 59.0 & 69.5 & 78.4 & 55.2 & 58.1 & 74.0 & \phantom{0}94.6 & 62.3 & 59.1 & 65.1 & 49.5 & 52.4 & 62.9  \\ 
  ~\citet{Tekin_2017_ICCV}\hypersetup{citecolor=black}  & n/a &54.2 &61.4 &60.2 &61.2& 79.4 &78.3& 63.1& 81.6& 70.1& 107.3& 69.3 &70.3 &74.3 &51.8& 63.2& 69.7\\
  ~\citet{sun2017compositional} \hypersetup{citecolor=black}& n/a& 52.8& 54.8& 54.2& 54.3 &61.8& 67.2& 53.1 &53.6& 71.7 &\phantom{0}86.7& 61.5& 53.4& 61.6& 47.1& 53.4& 59.1\\
 ~\citet{yang20183d}\hypersetup{citecolor=black}  & n/a&
  51.5 &58.9 &50.4& 57.0 &62.1 &65.4& 49.8 &52.7 &69.2 &\phantom{0}85.2 &57.4& 58.4 &\textbf{43.6}& 60.1& \underline{47.7}& 58.6\\
 ~\citet{hossain2018exploiting}\hypersetup{citecolor=black} & 16.9M & 48.4 & \underline{50.7} & 57.2 & 55.2 & 63.1 & 72.6 & 53.0 & 51.7 & 66.1 & \phantom{0}80.9 & 59.0 & 57.3 & 62.4 & 46.6 & 49.6 & 58.3  \\
 ~\citet{fang2018learning}\hypersetup{citecolor=black}  & n/a &  50.1& 54.3& 57.0 &57.1 &66.6& 73.3 &53.4& 55.7& 72.8& \phantom{0}88.6& 60.3& 57.7& 62.7 &47.5& 50.6& 60.4\\
 ~\citet{pavlakos2018ordinal}\hypersetup{citecolor=black}  & n/a & 48.5 & 54.4 & 54.5 & 52.0 & \underline{59.4} & 65.3 & 49.9 & 52.9 & \underline{65.8} & \phantom{0}\underline{71.1} & 56.6 & 52.9 & 60.9 & 44.7 & 47.8 & 56.2  \\
 SemGCN~\cite{zhaoCVPR19semantic}\hypersetup{citecolor=black} & 0.43M &  48.2& 60.8& 51.8& 64.0 &64.6& \textbf{53.6}& 51.1& 67.4& 88.7& \phantom{0}\textbf{57.7}& 73.2& 65.6& \underline{48.9}& 64.8& 51.9& 60.8\\
 ~\citet{sharma2019monocular}\hypersetup{citecolor=black}  & n/a & 48.6 & 54.5 & 54.2 & 55.7 & 62.2 & 72.0 & 50.5 & 54.3 & 70.0 & \phantom{0}78.3 & 58.1 & 55.4 & 61.4 & 45.2 & 49.7 & 58.0  \\ 
  GraphSH~\cite{xu2021graph}\hypersetup{citecolor=black} $\ast$ & 3.7M & \textbf{45.2} & \textbf{49.9} & \underline{47.5} & \underline{50.9} & \textbf{54.9} & 66.1 & 48.5 & \underline{46.3} & \textbf{59.7} &\phantom{0}71.5 & \textbf{51.4} & \textbf{48.6} & 53.9 & \textbf{39.9} & \textbf{44.1} & \textbf{51.9} \\ 
 Ours~(HG) & \underline{0.16M} & \underline{46.7} & 60.7 & \textbf{47.3} &\textbf{50.7} &64.1  &\underline{61.5} &\textbf{46.2}  &\textbf{45.3}  &67.1 &\phantom{0}80.4 & \underline{54.6} & \underline{51.4} & 55.4 & \underline{43.2} & 48.6 & \underline{52.1}  \\
 Ours~(HG) & \textbf{0.04M} &52.5& 61.4 & 47.8 & 53.0& 66.4 &65.4 & \underline{48.2} & \underline{46.3}  & 71.1& \phantom{0}84.3 & 57.8 & 52.3 & 57.0 & 45.7 &50.3  & 54.2  \\
 \hline
   \citet{liu2020comprehensive}~(GT) & 4.2M & 36.8 & \underline{40.3} & 33.0 & 36.3 & 37.5 & 45.0 & 39.7 & 34.9 & \underline{40.3} & \phantom{0}\underline{47.7} & 37.4 & 38.5 & 38.6 & 29.6 & \underline{32.0} & 37.8  \\ 
  GraphSH~\cite{xu2021graph}~(GT) & 3.7M & \underline{35.8} & \textbf{38.1} & \textbf{31.0} & \underline{35.3} & \textbf{35.8} & \textbf{43.2} & \underline{37.3} & \underline{31.7} & \textbf{38.4} & \phantom{0}\textbf{45.5} & \textbf{35.4} & \underline{36.7} & \textbf{36.8} & \textbf{27.9} & \textbf{30.7} & \textbf{35.8}  \\
    SemGCN~\cite{zhaoCVPR19semantic}~(GT) & 0.43M & 37.8 & 49.4 & 37.6 & 40.9 & 45.1 & 41.4 & 40.1 & 48.3 & 50.1 & \phantom{0}42.2 & 53.5 & 44.3 & 40.5 & 47.3 & 39.0 & 43.8  \\
 Ours~(GT) & \underline{0.16M} & \textbf{31.2} & 46.9 & \underline{32.5} & \textbf{31.7} & \underline{41.4} & \underline{44.9} & \textbf{33.9} & \textbf{30.9}& 49.2 & \phantom{0}55.7 &\underline{35.9} &\textbf{36.1} & \underline{37.5} &  \underline{29.07} & 33.1& \underline{36.2}  \\ 
 Ours~(GT) & \textbf{0.04M} & 33.6 & 48.5  & 34.9 &34.8 & 46.0 & 49.5 & 36.7 & 33.7 & 50.6 & \phantom{0}62.7 & 38.9 & 40.3 & 41.4 & 33.1 & 36.3 & 40.0 \\ 
\hline
 \end{tabular}
 }

 \caption{
 Quantitative comparisons w.r.t. MPJPE (in mm) on Human3.6M~\cite{h36m_pami} under Protocol~\#1. Best in bold, second-best underlined.
 In the upper part, all methods use stacked hourglass (HG) 2D estimates~\cite{newell2016stacked} as inputs, except for~\cite{xu2021graph} (which uses CPN~\cite{Chen_2018_CVPR}, indicated by $\ast$).
 In the lower part, all methods use the 2D ground truth (GT) as input.
 }
 \label{protocol1}
\end{table*}
\vspace{-30px}
\section{Experimental Results}
\subsection{Datasets and Evaluation Protocols}
We use the publicly available motion capture dataset Human3.6M~\cite{h36m_pami}. It contains 3.6 million images produced by 11 actors performing 15 actions. Four different calibrated RGB cameras are used to capture the subjects during training and test time. Same as previous works, \eg~\cite{martinez2017simple, Tekin_2017_ICCV, sun2017compositional, pavlakos2018ordinal, zhaoCVPR19semantic, sharma2019monocular, xu2021graph}, we use five subjects (S1, S5, S6, S7, S8) for training and two subjects (S9 and S11) for testing. Each sample from the  different camera views is considered independently. We also use MPI-INF-3DHP dataset~\cite{mono-3dhp2017} to test the generalizability of our model. MPI-INF-3DHP contains 6 subjects for testing in three different scenarios: studio with a green screen~(GS), studio without green screen~(noGS), and outdoor scene~(Outdoor). Note that for experiments on MPI-INF-3DHP we also only trained on Human3.6M.

Following~\cite{martinez2017simple, Tekin_2017_ICCV, sun2017compositional, zhaoCVPR19semantic, xu2021graph}, we use the MPJPE protocol, referred to as Protocol~\#1. MPJPE is the mean per joint position error in millimeters between predicted joint positions and ground truth joint positions after aligning the pre-defined root joints (\ie the pelvis joint). Note that some works~(\eg~\cite{pavllo20193d, liu2020comprehensive}) use the P-MPJPE metric, which reports the error after a rigid transformation to align the predictions with the ground truth joints. We explicitly select the standard MPJPE metric as it is more challenging and also allows for a fair comparison to previous related works. For the MPI-INF-3DHP test set, similar to previous works~\cite{xu2021graph, luo2018}, we use the percentage of correct 3D keypoints (3D PCK) within a 150 mm radius~\cite{mono-3dhp2017} as evaluation metric.
\vspace{-10px}
\subsection{Implementation Details}
\textbf{2D Pose Estimation.} The inputs to our architecture are the 2D joint positions estimated from the RGB images for all four cameras independently. Our method is independent of the off-the-shelf architecture used for estimating 2D joint positions. Similar to previous works~\cite{martinez2017simple, zhaoCVPR19semantic}, we use the stacked hourglass architecture~\cite{newell2016stacked} to estimate the 2D joint positions. The hourglass architecture is an autoencoder architecture that stacks the encoder-decoder with skip connections multiple times. Following~\cite{zhaoCVPR19semantic}, the stacked hourglass network is first pre-trained on the MPII~\cite{andriluka14cvpr} dataset and then fine-tuned on the Human3.6M~\cite{h36m_pami} dataset. As described in~\cite{pavllo20193d}, the input joints are scaled to image coordinates and normalized to $[-1, 1]$.

\textbf{3D Pose Estimation.} The ground truth 3D joint positions in the Human3.6M dataset are given in world coordinates. Following previous works~\cite{zhaoCVPR19semantic, martinez2017simple}, we transform the joint positions to the camera space given the camera calibration parameters.
Similar to previous works~\cite{zhaoCVPR19semantic, martinez2017simple}, to make the architecture trainable, we chose a predefined joint (the pelvis joint) as the center of the coordinate system. We do not use any augmentations throughout all our experiments.

We trained our architecture using Adam~\cite{kingma2014adam} with an initial learning rate of $0.001$ and used mini-batches of size $64$. The learning rate is dropped with a decay rate of $0.5$ when the loss on the validation set saturates. The architecture contains seven M\"obiusGCN blocks, where each block, except the first and the last block with the input and the output channels $2$ and $3$ respectively, contains either $64$ channels (leading to $0.04\text{M}$ parameters) or $128$ channels (leading to $0.16\text{M}$ parameters). We initialized the weights using the Xavier method~\cite{glorot2010understanding}. 
During the test phase, the scale of the outputs is calibrated by forcing the sum of the length of all 3D bones to be equal to a canonical skeleton~\cite{pavlakos2017coarse, zhou2017towards, zhou2018monocap}. To help the architecture differentiate between different 3D poses with the same 2D pose, similar to~\citet{poier2018learning}, we provide the center of mass of the subject to the architecture as an additional input. Same as~\cite{martinez2017simple, zhaoCVPR19semantic}, we predict $16$ joints (\ie without the 'Neck/Nose' joint).

Also, as in previous works~\cite{martinez2017simple, zhaoCVPR19semantic, pavlakos2017coarse, liu2020comprehensive}, our network predicts the normalized locations of 3D joints. We did all our experiments on an NVIDIA GeForce RTX 2080 GPU using the PyTorch framework~\cite{NEURIPS2019_9015}.
For the loss function, same as previous works \emph{\eg}~\cite{martinez2017simple, pavllo20193d}, we use the mean squared error (MSE) between the 3D ground truth joint locations $\mathcal{Y}$ and our predictions $\mathcal{\hat{Y}}$, \ie
\begin{equation}
    \mathcal{L}(\mathcal{Y}, \mathcal{\hat{Y}}) = \sum_{i=1}^{\kappa}(\mathcal{Y}_i-\mathcal{\hat{Y}}_i)^2.
\end{equation}

\textbf{Complex-valued M\"obiusGCN.} In complex-valued neural networks, the data and the weights are represented in the complex domain. A complex function is holomorphic (complex-differentiable) if not only their partial derivatives exist but also satisfy the Cauchy-Riemann equations. Complex neural networks have different applications, \eg~\citet{wolter2018complex} proposed a complex-valued recurrent neural network which helps solving the exploding/vanishing gradients problem in RNNs. 
Complex-valued neural networks are easier to optimize than real-valued neural networks and have richer representational capacity~\cite{trabelsi2018deep}. Considering the Liouville theorem~\cite{mandic2009complex}, designing a fully complex differentiable (holomorphic) neural network is hard as only constant functions are both holomorph and bounded. Nevertheless, it was shown that in practice full complex differentiability of complex neural networks is not necessary~\cite{trabelsi2018deep}.

In complex-valued neural networks, the complex convolution operator is defined as 
\begin{equation*}
    \mathbf{W} \ast \mathbf{h} = (\mathbf{A} \ast \mathbf{x} - \mathbf{B} \ast \mathbf{y}) + i(\mathbf{B} \ast \mathbf{x} + \mathbf{A} \ast \mathbf{y}),
\end{equation*}
where $\mathbf{W} = \mathbf{A} + i\mathbf{B}$ and $\mathbf{h} = \mathbf{x} + i\mathbf{y}$. $\mathbf{A}$ and $\mathbf{B}$ are real matrices and $\mathbf{x}$ and $\mathbf{y}$ are real vectors. 
We also apply the same operators on our graph signals and graph filters.
The PyTorch framework~\cite{NEURIPS2019_9015} utilizes Wirtinger calculus~\cite{kreutz2009complex} for backpropagation, which  optimizes the real and imaginary partial derivatives independently.

\begin{figure*}
\centering
\subfloat[]{\includegraphics[width=0.3\textwidth]{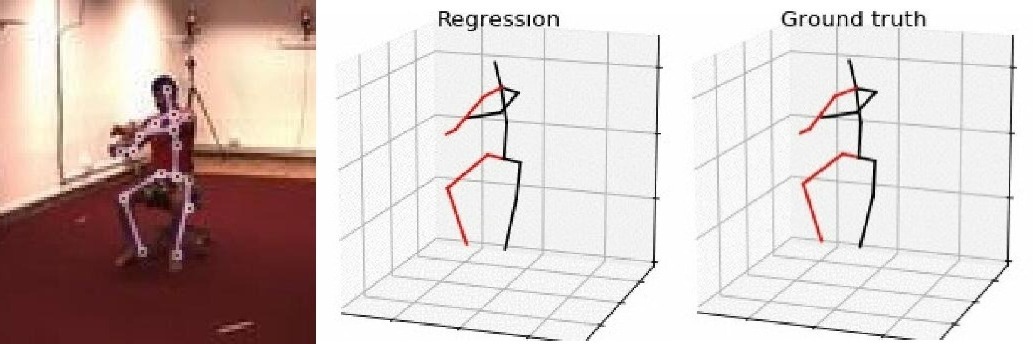}}
\quad
\subfloat[]{\includegraphics[width=0.3\textwidth]{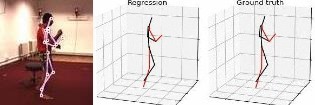}}
\quad
\subfloat[]{\includegraphics[width=0.3\textwidth]{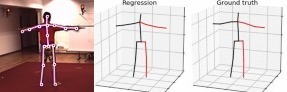}}\\
\subfloat[]{\includegraphics[width=0.3\textwidth]{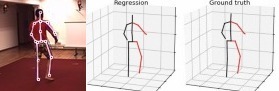}}
\quad
\subfloat[]{\includegraphics[width=0.3\textwidth]{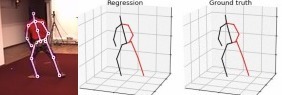}}
\quad
\subfloat[]{\includegraphics[width=0.3\textwidth]{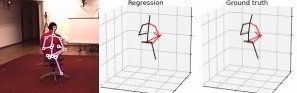}}\\
\subfloat[]{\includegraphics[width=0.3\textwidth]{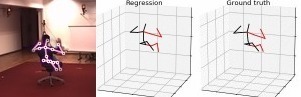}}
\quad
\subfloat[]{\includegraphics[width=0.3\textwidth]{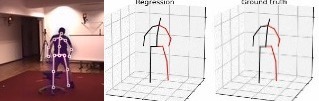}}
\quad
\subfloat[]{\includegraphics[width=0.3\textwidth]{qual/images11.jpg}}\\
\caption{Qualitative results of M\"obiusGCN on Human3.6M~\cite{h36m_pami}.}
\label{qual-res}
\end{figure*}

\vspace{-20px}
\subsection{ Fully-supervised M\"{o}biusGCN}
In the following, we compare the results of M\"obiusGCN in a fully-supervised setup with the previous state-of-the-art for 3D human pose estimation on the Human3.6M and MPI-INF-3DHP datasets. For this, we use a)~estimated 2D poses using the stacked hourglass architecture (HG)~\cite{newell2016stacked} as input and b)~the 2D ground truth (GT).

\textbf{Comparisons on Human3.6M.}
Table~\ref{protocol1} shows the comparison of our M\"obiusGCN to the state-of-the-art methods under Protocol~\#1 on Human3.6M dataset. 

By setting the number of channels to $128$ in each block of the M\"obiusGCN ($0.16\text{M}$ parameters), given estimated 2D joint positions (HG), we achieve an average MPJPE of $52.1\text{mm}$ over all actions and test subjects. Using the ground truth (GT) 2D joint positions as input, we achieve an MPJPE of $36.2\text{mm}$.
These results are on par with the state-of-the-art, \ie~GraphSH~\cite{xu2021graph}, which achieves $51.9\text{mm}$ and $35.8\text{mm}$, respectively.
Note, however, that M\"obiusGCN drastically reduces the number of training parameters by up to $96\%$ ($0.16\text{M}$ \emph{vs.} $3.7\text{M}$).

\begin{figure}
\centering
\subfloat{\includegraphics[width=0.3\textwidth]{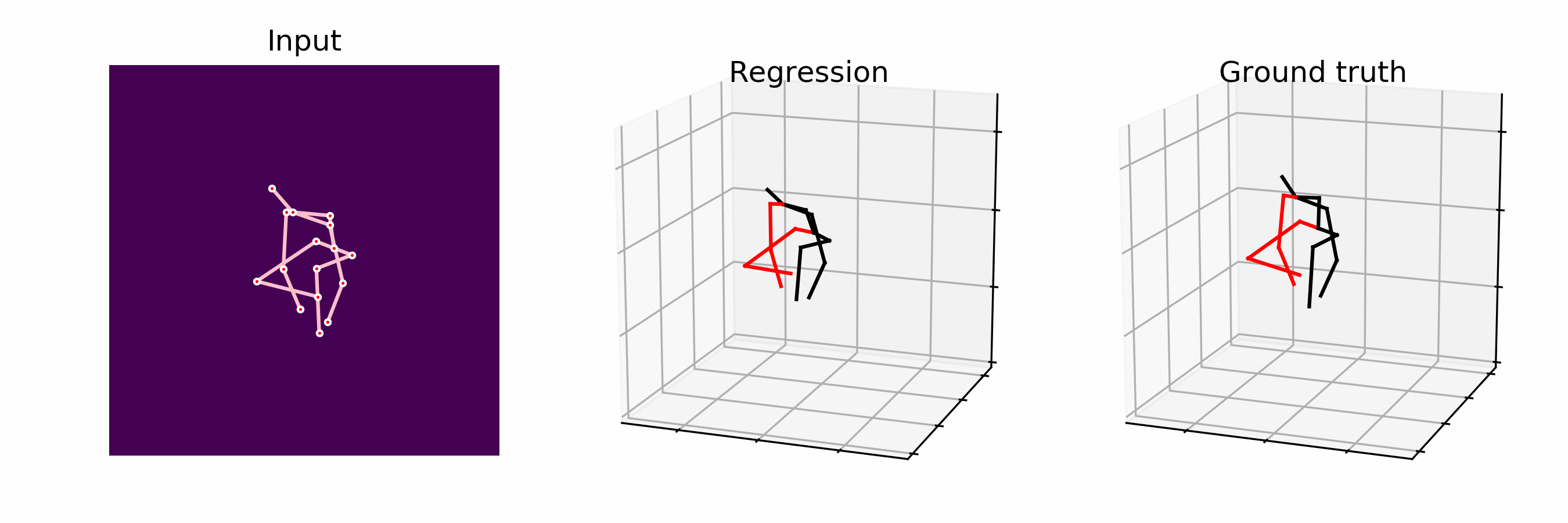}}
\quad
\subfloat{\includegraphics[width=0.3\textwidth]{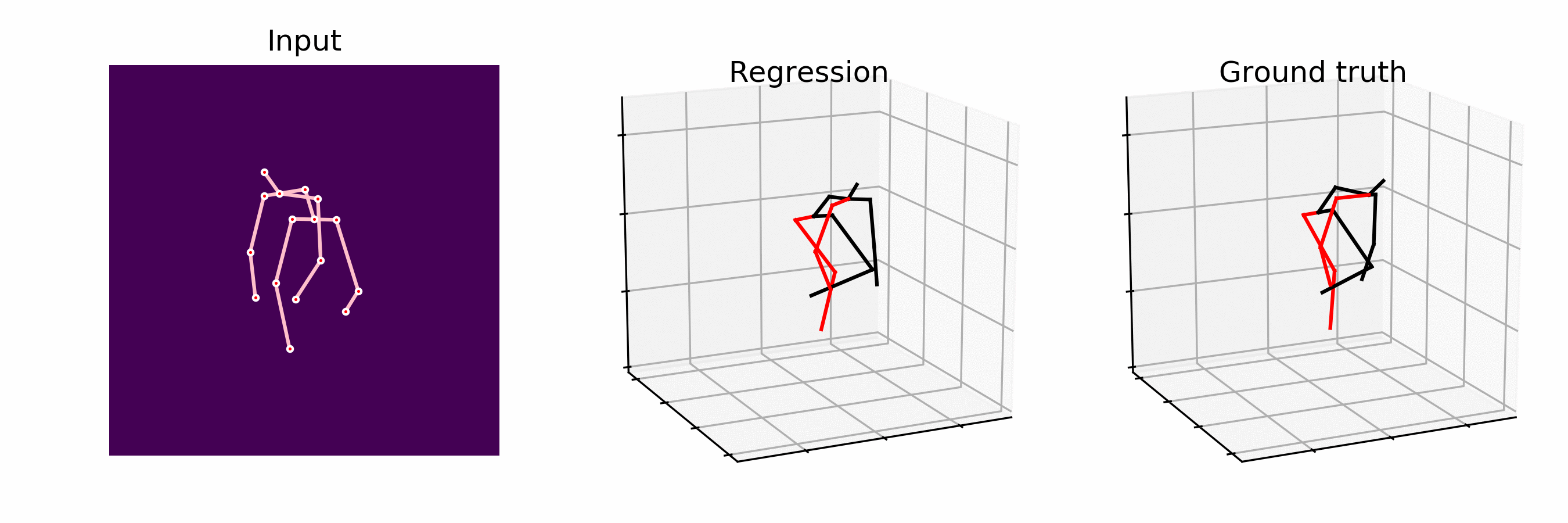}}
\quad
\subfloat{\includegraphics[width=0.3\textwidth]{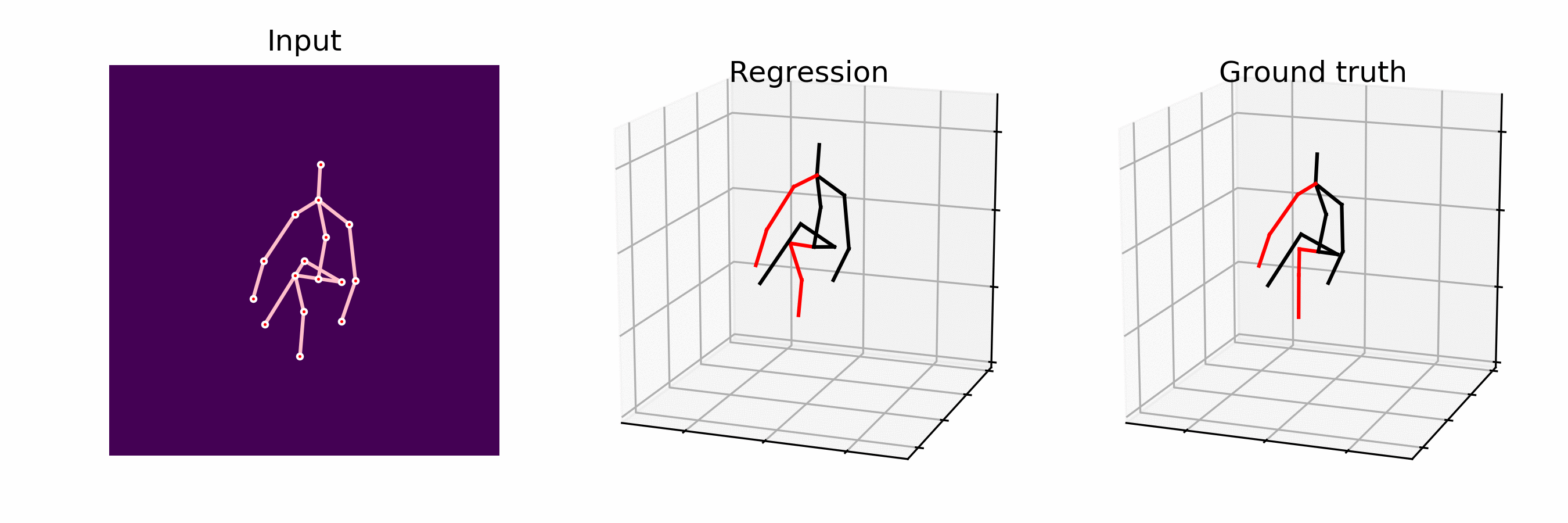}}\\ 
\subfloat{\includegraphics[width=0.3\textwidth]{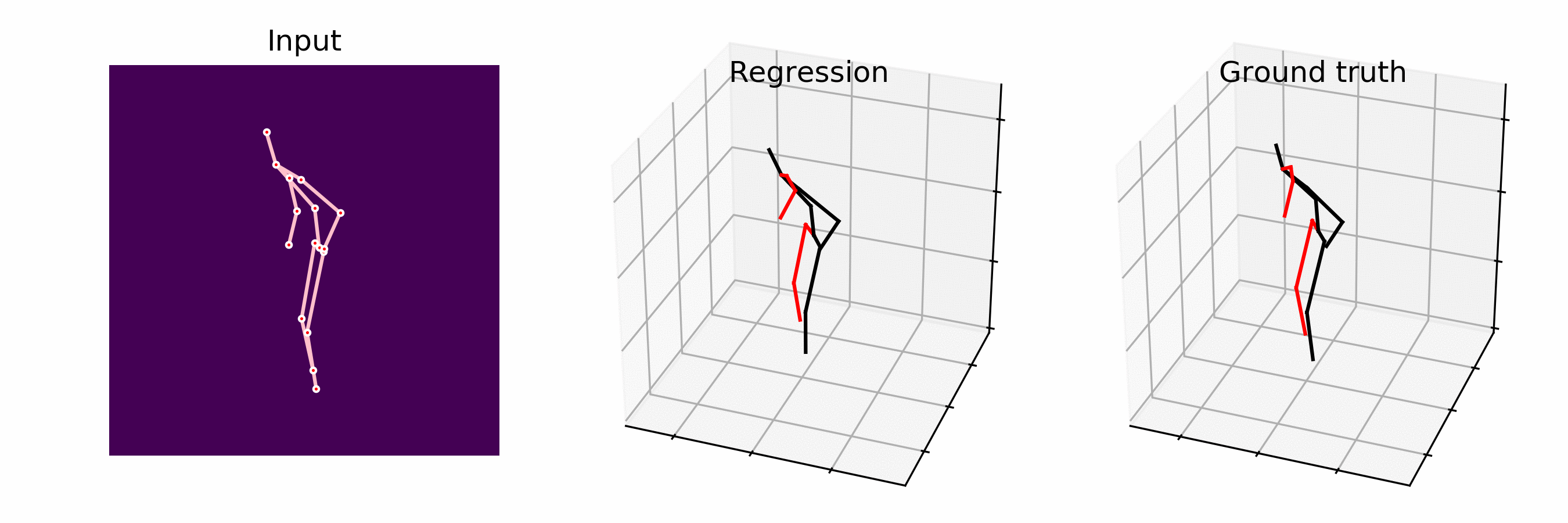}}
\quad
\subfloat{\includegraphics[width=0.3\textwidth]{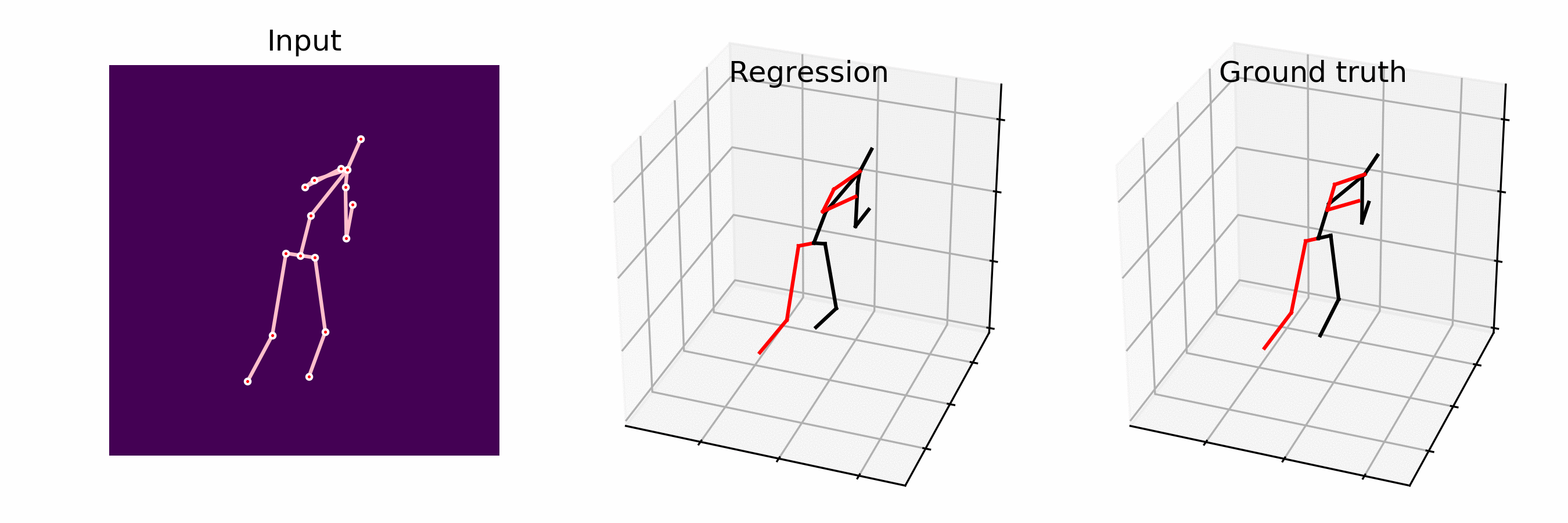}}
\quad
\subfloat{\includegraphics[width=0.3\textwidth]{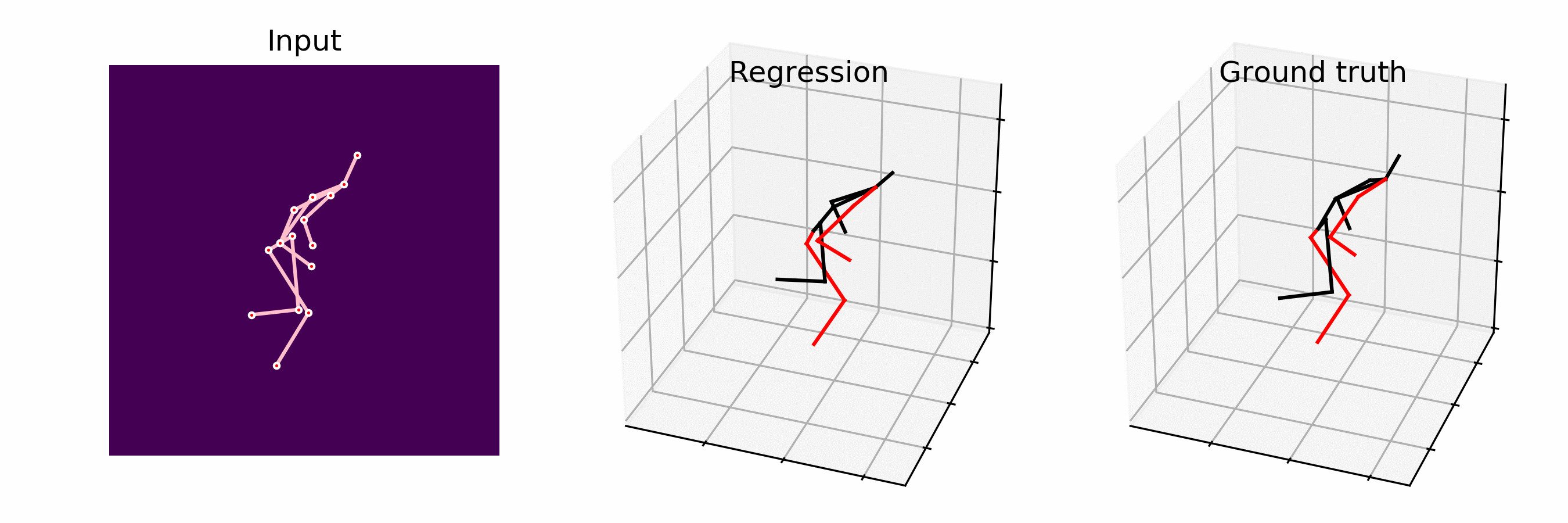}} \\
\subfloat{\includegraphics[width=0.3\textwidth]{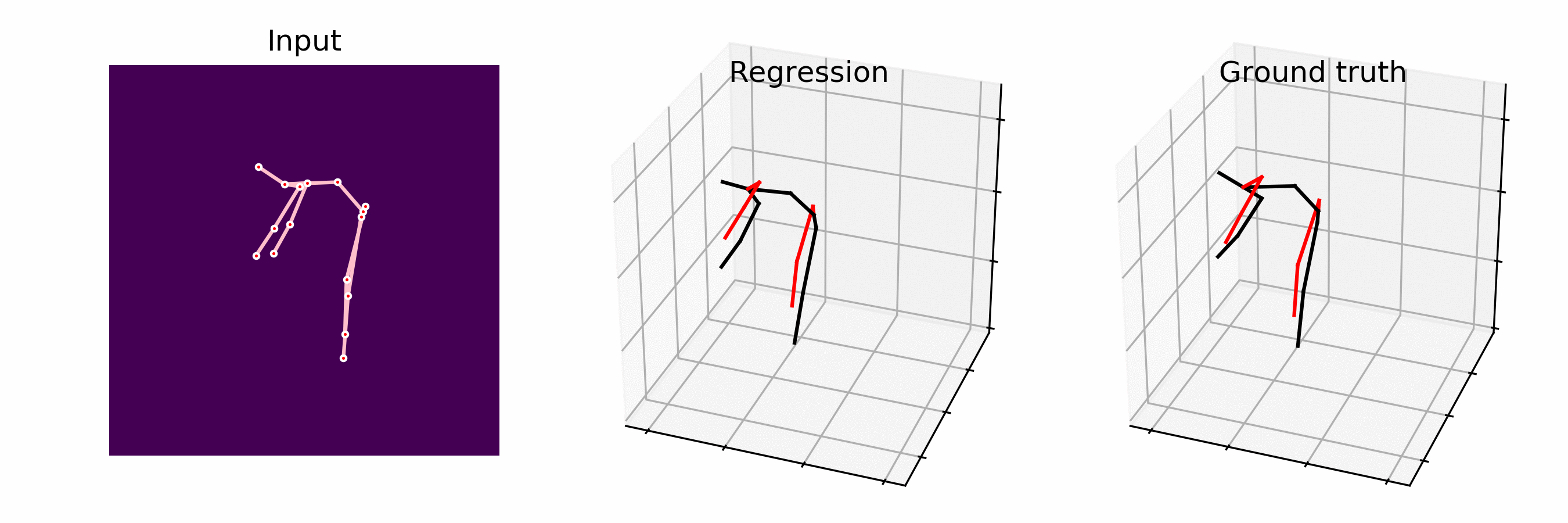}}
\quad
\subfloat{\includegraphics[width=0.3\textwidth]{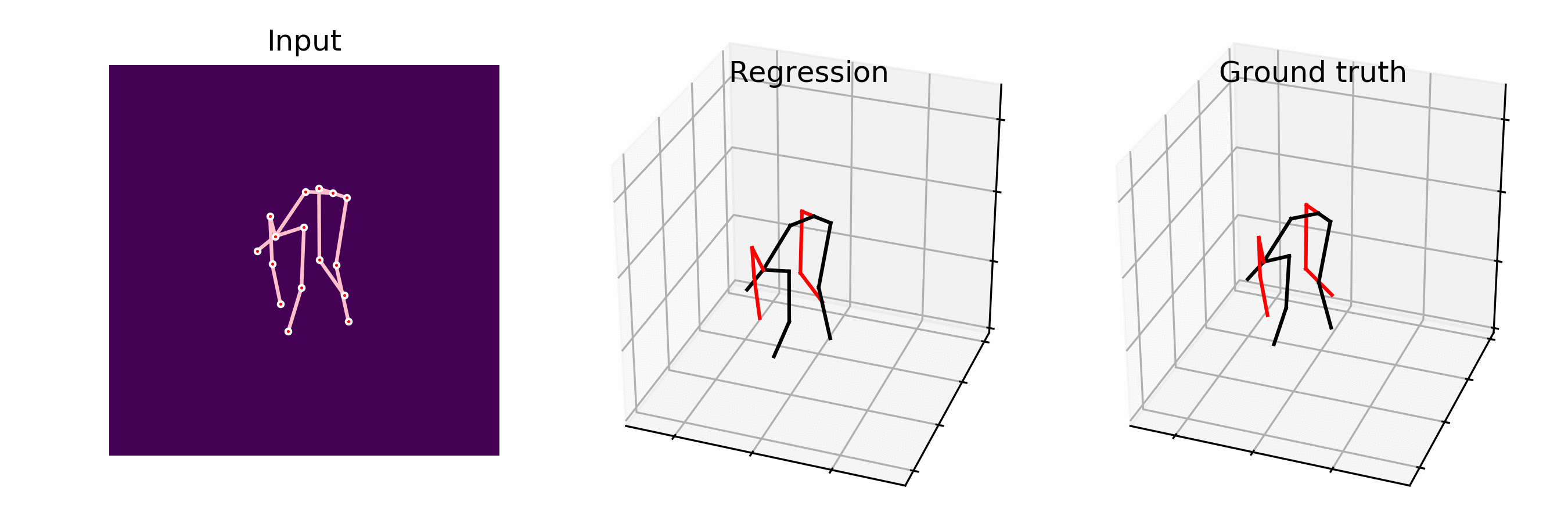}}
\quad
\subfloat{\includegraphics[width=0.3\textwidth]{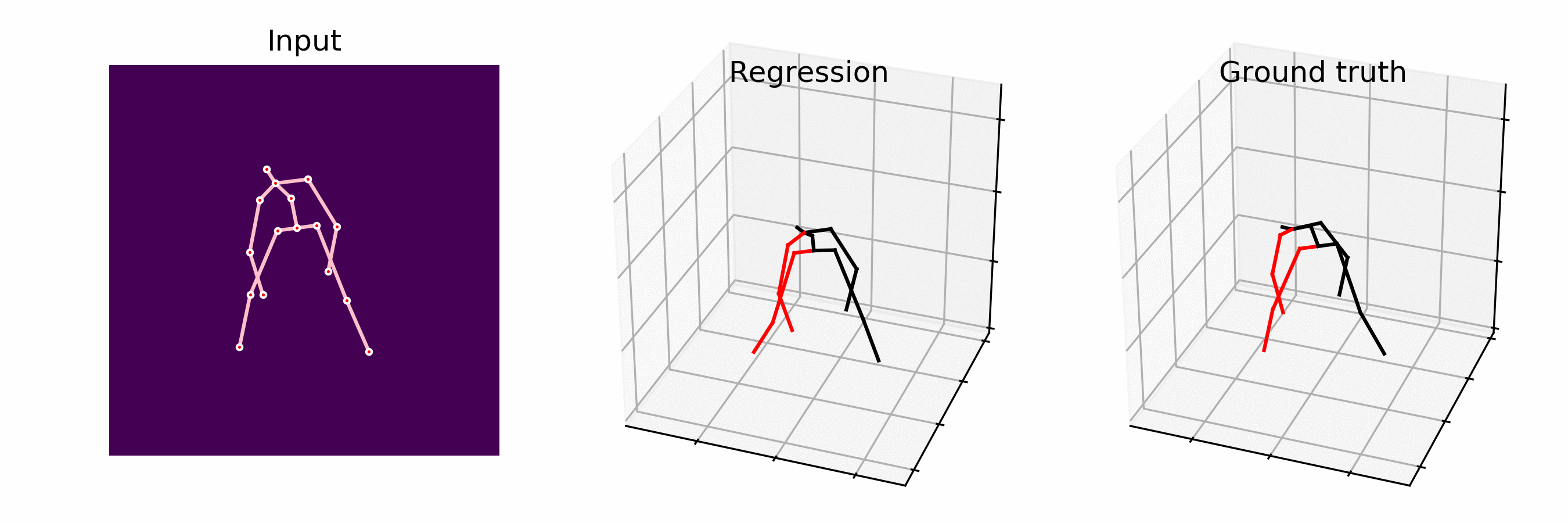}} \\
\caption{\scriptsize{Qualitative self-occlusion results of M\"obiusGCN on MPI-INF-3DHP~\cite{mono-3dhp2017} (trained only on Human3.6m).}}
\label{qual-res-mpi-inf}
\end{figure}

Reducing the number of channels to $64$, we still achieve impressive results considering the lightness of our architecture (\ie~only $0.042\text{M}$ parameters).
Compared to GraphSH~\cite{xu2021graph}, we reduce the number of parameters by $98.9\%$ ($0.042\text{M}$ vs $3.7\text{M}$) and still achieve notable results, \ie~MPJPE of $40.0\text{mm}$ \emph{vs.} $35.8\text{mm}$ (using 2D GT inputs) and $54.2\text{mm}$ \emph{vs.} $51.9\text{mm}$ (using the 2D HG inputs).
Note that GraphSH~\cite{xu2021graph} is the only approach which uses a better 2D pose estimator as input (CPN~\cite{Chen_2018_CVPR} instead of HG~\cite{newell2016stacked}). Nevertheless, our M\"obiusGCN (with 0.16M parameters) achieves competitive results.
Furthermore, M\"obiusGCN outperforms the previously lightest architecture SemGCN~\cite{zhaoCVPR19semantic}, \ie~$40.0\text{mm}$ vs $43.8\text{mm}$ (using 2D GT inputs) and $54.2\text{mm}$ vs $60.8\text{mm}$ (using 2D HG input), although we require only $9.7\%$ of their number of parameters ($0.042\text{M}$ \emph{vs.} $0.43\text{M}$).
Figure~\ref{qual-res} shows qualitative results of our M\"obiusGCN with $0.16\text{M}$ parameters on unseen subjects of the Human3.6M dataset given the 2D ground truth (GT) as input.
\vspace{-25px}
\begin{center}
\resizebox{5.5CM}{!}{%
\begin{tabular}{c| c c c c c}
\toprule
Method & \# Parameters & GS & noGS & Outdoor & All(PCK) \\
\midrule
\citet{martinez2017simple} & 4.2M & 49.8 & 42.5 & 31.2 & 42.5 \\
\citet{mono-3dhp2017} & n/a & 70.8 & 62.3 & 58.8 & 64.7 \\
\citet{luo2018} & n/a & 71.3& 59.4& 65.7& 65.6 \\
\citet{yang20183d} & n/a & - & - & - & 69.0 \\
\citet{zhou2017towards} & n/a & 71.1 & 64.7 & 72.7 & 69.2 \\
\citet{ci2019optimizing} & n/a & 74.8& 70.8& 77.3 &74.0 \\
\citet{zhou2019hemlets} & n/a & 75.6& 71.3& 80.3 &75.3 \\
GraphSH~\cite{xu2021graph} & 3.7M & \textbf{81.5} & \textbf{81.7} & \underline{75.2} & \textbf{80.1} \\
\midrule
Ours & \textbf{0.16M} & \underline{79.2} & \underline{77.3} & \textbf{83.1} & \underline{80.0} \\ 
\bottomrule
\end{tabular}
} \captionof{table}{Results on the MPI-INF-3DHP test set~\cite{mono-3dhp2017}. Best in bold, second-best underlined.}
\label{mpi-inf-3dhp}
 \end{center}
 \vspace{-25px}
\textbf{Comparisons on MPI-INF-3DHP.} The quantitative results on MPI-INF-3DHP~\cite{mono-3dhp2017} are shown in~\autoref{mpi-inf-3dhp}. Although we train M\"obiusGCN only on the Human3.6M~\cite{h36m_pami} dataset and our architecture is lightweight, the results indicate our strong generalization capabilities to unseen datasets, especially for the most challenging outdoor scenario.
Figure~\ref{qual-res-mpi-inf} shows some qualitative results on unseen self-occlusion examples from the test set of MPI-INF-3DHP dataset with M\"obiusGCN trained only on Human3.6M. 

Though M\"obiusGCN for 3D human pose estimation has comparably fewer parameters, it is computationally expensive, \ie~$\mathcal{O}(n^3)$, both in the forward and backward pass due to the decomposition of the Laplacian matrix. In practice, however, this is not a concern for human pose estimation because of the small human pose graphs, \ie$\sim20$ nodes. 
More specifically, a single forward pass takes on average only $0.001\,\text{s}$.
\begin{center}
\resizebox{4CM}{!}{%
\begin{tabular}{c|c c}
\toprule
 Method & \# Parameters & MPJPE  \\
\midrule
\citet{liu2020comprehensive}\hypersetup{citecolor=black} & 4.20M & \phantom{0}37.8 \\
GraphSH~\cite{xu2021graph}\hypersetup{citecolor=black} & 3.70M & \phantom{0}\textbf{35.8} \\
\citet{liu2020comprehensive}\hypersetup{citecolor=black} & 1.05M & \phantom{0}40.1 \\
GraphSH~\cite{xu2021graph}\hypersetup{citecolor=black} & 0.44M & \phantom{0}39.2 \\
SemGCN~\cite{zhaoCVPR19semantic}\hypersetup{citecolor=black} & 0.43M & \phantom{0}43.8 \\
\citet{yan2018spatial}\hypersetup{citecolor=black} & 0.27M & \phantom{0}57.4 \\
\citet{velivckovic2017graph}\hypersetup{citecolor=black} & \underline{0.16M} & \phantom{0}82.9 \\
Chebychev-GCN & 0.08M & 110.6\\
\midrule
Ours  & \underline{0.16M} & \phantom{0}\underline{36.2} \\
Ours  & \textbf{0.04M} & \phantom{0}40.0\\
\bottomrule
\end{tabular}
}
 \captionof{table}{Supervised quantitative comparison between GCN architectures on Human3.6M~\cite{h36m_pami} under Protocol~\#1. Best in bold, second-best underlined. All methods use 2D ground truth as input.}
\label{light-archs}
 \end{center}
\textbf{Comparison to Previous GCNs.}
Table~\ref{light-archs} shows our performance in comparison to previous GCN architectures.
Besides significantly reducing the number of required parameters, applying the M\"obius transformation also allows us to leverage better feature representations.
Thus, M\"obiusGCN can outperform all other light-weight GCN architectures.
It even achieves better results ($36.2\text{mm}$ \emph{vs.} $39.2\text{mm}$) than the light-weight version of the state-of-the-art GraphSH~\cite{xu2021graph}, which requires $0.44\text{M}$ parameters. 

We also compare our proposed spectral GCN with the vanilla spectral GCN, \ie Chebychev-GCN~\cite{kipf2016semi}. Each block of Chebychev-GCN is the real-valued spectral GCN from~\cite{kipf2016semi}. We use $7$ blocks, similar to our M\"obiusGCN, with $128$ channels each. Our complex-valued M\"obiusGCN with only $0.04\text{M}$ clearly outperforms the Chebychev-GCN~\cite{kipf2016semi} with $0.08\text{M}$ parameters ($40.0\text{mm}$ \emph{vs.} $110.6\text{mm}$). This highlights the representational power of our M\"obiusGCN in contrast to vanilla spectral GCNs.
\vspace{-10px}
\begin{center}
\resizebox{4CM}{!}{
 \begin{tabular}{c| c c c c } 
 \toprule
 Method & Temp &MV &Input & MPJPE \\ [0.5ex] 
 \midrule
 \citet{rhodin2018unsupervised}\hypersetup{citecolor=black} &\xmark  & \checkmark  &  RGB  & 131.7 \\
\citet{pavlakos2019texturepose}\hypersetup{citecolor=black}&\checkmark  &\checkmark  &RGB & 110.7 \\
\citet{chen2019weakly}\hypersetup{citecolor=black}&\xmark  &\checkmark  &  HG & \phantom{0}91.9 \\
\citet{li2019boosting}\hypersetup{citecolor=black} &\checkmark  &\xmark  &  RGB & \phantom{0}\underline{88.8}  \\
Ours (0.16M) &\xmark  &\xmark  &  HG & \phantom{0}\textbf{82.3}\\
\midrule
\citet{iqbal2020weakly}\hypersetup{citecolor=black}&\xmark &\checkmark & GT & \phantom{0}\underline{62.8} \\
 Ours (0.16M) &\xmark  &\xmark  &  GT & \phantom{0}\textbf{62.3}\\  
 \bottomrule
 \end{tabular}
} \captionof{table}{ Semi-supervised quantitative comparison on Human3.6M~\cite{h36m_pami} under Protocol~\#1. Temp, MV, GT, and HG stand for temporal, multi-view, ground-truth, and stacked hourglass as 2D pose input respectively. Best in bold, second-best underlined.}
\label{semi-supervised}
 \end{center}
\vspace{-10px}
\subsection{M\"obiusGCN with Reduced Dataset}
A major practical limitation with training neural network architectures is to acquire sufficiently large and accurately labeled datasets. Semi-supervised methods try to address this by combining fewer labeled samples with large amounts of unlabeled data.
Another benefit of M\"obiusGCN is that we require fewer training samples. Having a better feature representation in M\"obiusGCN leads to a light architecture and therefore, requires less training samples. 

To demonstrate this, we train M\"obiusGCN with a limited number of samples. In particular, we use only one subject to train M\"obiusGCN and do not need any unlabeled data.
Table~\ref{semi-supervised} compares the M\"obiusGCN to the semi-supervised approaches~\cite{rhodin2018unsupervised, chen2019weakly,pavlakos2019texturepose,li2019boosting, iqbal2020weakly}, which were trained using both labeled and unlabeled data.
As can be seen, M\"obiusGCN performs favorably: we achieve an MPJPE of $82.3 \text{mm}$ (given 2D HG inputs) and an MPJPE of $62.3 \text{mm}$ (using the 2D GT input). In contrast to previous works, we neither utilize other subjects as weak supervision nor need large unlabeled datasets during training. 

As shown in Table~\ref{semi-supervised}, M\"obiusGCN also outperforms methods which rely on multi-view cues~\cite{rhodin2018unsupervised, chen2019weakly} or leverage temporal information~\cite{li2019boosting}.
Additionally, we achieve better results to~\cite{iqbal2020weakly}, even though, in contrast to this approach, we do not incorporate multi-view information or require extensive amounts of unlabeled data during training.

Table~\ref{semisupervised-archs} analyzes the effect of increasing the number of training samples.
As can be seen, our M\"obiusGCN only needs to train on three subjects to perform on par with SemGCN~\cite{zhaoCVPR19semantic}.
\vspace{-5px}
\begin{center}
\resizebox{5CM}{!}{%
 \begin{tabular}{c| c c c} 
\toprule
 Method & Subject & \# Parameters & MPJPE \\ [0.5ex] 
 \midrule
 Ours & S1 &  0.16M & 62.3 \\
Ours & S1 S5 & 0.16M & 47.9 \\
Ours & S1 S5 S6 & \textbf{0.16M} & \textbf{43.1} \\
\midrule
 SemGCN~\cite{zhaoCVPR19semantic}& All subjects & 0.43M & 43.8 \\\bottomrule
 \end{tabular}
}
 \captionof{table}{Evaluating the effects of using fewer training subjects on Human3.6M~\cite{h36m_pami} under Protocol~\#1 (given 2D GT inputs).}
\label{semisupervised-archs}
 \end{center}
\vspace{-17px}
\section{Conclusion and Discussion}
\vspace{-5px}
We proposed a novel rational spectral GCN (M\"obiusGCN) to predict 3D human pose estimation by encoding the transformation between joints of the human body, given the human body joint positions in 2D. Our proposed method achieves state-of-the-art result accuracy while preserving the compactness of the model with lower number of parameters than the most compact model existing in the literature (lower number of parameters by an order of magnitude).
We verified the generalizability of our model on the MPI-INF-3DHP dataset, where we achieve state-of-the-art results on the most challenging in-the-wild (outdoor) scenario.

Our proposed simple and light-weight architecture requires less data for training. This allows us to outperform the previous lightest architecture by just training our model with three subjects on the Human3.6M dataset. We also showed promising results of our architecture in comparison to previous state-of-the-art semi-supervised architectures despite not using any temporal or multi-view information or large unlabeled datasets.

\textbf{Acknowledgement.} This research was funded by the Austrian Research Promotion Agency (FFG) under project no. 874065 and the ERC grant no. 802554 (SPECGEO).
\renewcommand{\bibname}{\leftline{References}}
\bibliographystyle{plainnat}
\bibliography{egbib}
\end{document}